# Model-based thermal drift compensation for high-precision hexapod robot actuators

Clément ROBERT [1, 2] Alain VISSIERE [3] Olivier COMPANY [1] Pierre NOIRE [2] Thierry ROUX [2] and Sébastien KRUT [1]

[1] LIRMM (Laboratoire d'Informatique, de Robotique et de Microélectronique de Montpellier), UMR5506 Montpellier Univ. and CNRS, Montpellier, France.
[2] SYMETRIE, Nîmes, France.
[3] LCM (Laboratoire Commun de Métrologie), LNE/CNAM, Paris, France.

*Submitting author e-mail address: clement.robert@lirmm.fr*

**Abstract**
Thermal expansion is a significant source of positioning error in high-precision hexapod robots (Gough-Stewart platforms). Any variation in the temperature of the hexapod's parts induces expansion, which alters their kinematic model and reduces the robot's accuracy and repeatability. These variations may arise from internal heat sources (such as motors, encoders, and electronics) or from environmental changes. In this study, a method is proposed to anticipate and therefore correct the thermal drift of one of the hexapod precision electro-mechanical actuators. This method is based on determining a model that links the expansion state of the actuator at any given moment to the temperature of some well-chosen points on its surface. This model was initially developed theoretically. Its coefficients were then adjusted experimentally on a specific test-bench, based on a rigorous measurement campaign of actuator expansion using a high-precision interferometric measurement system. Experimental validation demonstrates a reduction of thermally induced expansion by more than 80%. This paves the way for thermal drift correction across the entire robot or similar robotics parts.

**1. Introduction**

Hexapod robots [1] are parallel robots with six degrees of freedom (DoFs), typically classified as motion hexapods (fast-moving [2,3]) or positioning hexapods (high accuracy and resolution). Positioning hexapods are increasingly deployed in industrial settings, including aerospace and automotive production lines [4–6]. Optimal performance requires both stability and repeatability [7-8], yet various sources of error can degrade these qualities. Among them, thermal expansion during operation—thermal drift [9]—is particularly significant, arising from external (ambient temperature) and internal (motor heating) sources. This study addresses the correction of thermal drift in a precision positioning hexapod with a 30,000 cm³ workspace and 250 mm leg stroke.

While thermal effects have been extensively studied in machine tools [10–16], they remain underexplored in industrial robots due to historically lower precision. However, robot accuracy has increased 10–100× since the 1970s [17], making thermal deformation a major error source. Prior research on serial robots [18-19] and limited parallel structures [17,20] has shown partial compensation is possible, but residual errors often exceed tens of micrometres—far above the 1–2 µm target of this work.

Thermal drift mitigation generally follows three strategies: temperature control, structural design, or compensation [21–23]. Active environmental control is impractical in industrial hexapod applications, and material selection alone cannot eliminate thermal effects. Compensation approaches are therefore required. Direct methods periodically measure drift and adjust the setpoint [24,25], but this interrupts robot operation. Indirect methods estimate drift via a continuous mathematical model relating thermal expansion to measurable quantities, usually temperature, often using regression analysis to derive model coefficients [26–28].

This study focuses on a single hexapod leg, as it exhibits the largest thermal errors, while other components contribute minimally. The objectives are to (i) establish a regression-based thermal correction model, (ii) determine optimal temperature sensor locations, and (iii) experimentally validate the model. The approach entails:

1. Installing temperature sensors and a thermal expansion measurement system on one leg.
2. Recording leg thermal expansion and temperature during operation.
3. Developing a linear regression model linking thermal deformation to temperature.
4. Validating the model through additional experiments.

This methodology enables precise thermal drift compensation for high-accuracy parallel robots, with potential extension to complete hexapod systems.

## 2. Presentation of one of the hexapod legs

The CAD model of one leg is shown in Fig. 1.

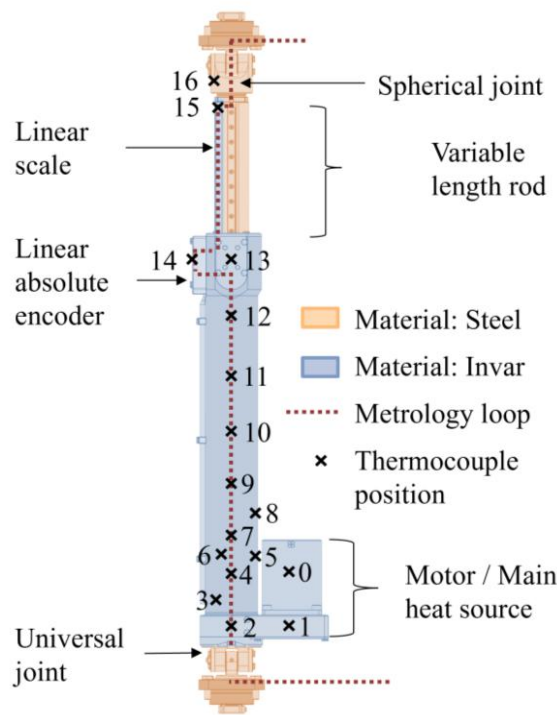

**Figure 1.** Complete leg CAD model.

The motor drives mechanical elements in the lower part, transmitting motion to a screw inside the fixed section, which in turn moves a sliding rod. A linear scale mounted on the rod measures its displacement with high accuracy. To reach the commanded position, the control software corrects for known mechanical errors (e.g., defects, backlash), but not for thermal expansion. The corrected setpoint thus yields the actual position $q_{actual}$, ideally equal to $q_{setpoint}$. However, discrepancies remain. Previous studies [23] identified thermal expansion as the dominant residual error source, leading to the assumption that the remaining deviation, $\Delta q_{actual}^{thermal}$, arises solely from thermal drift. The notations are summarized in Fig. 2.

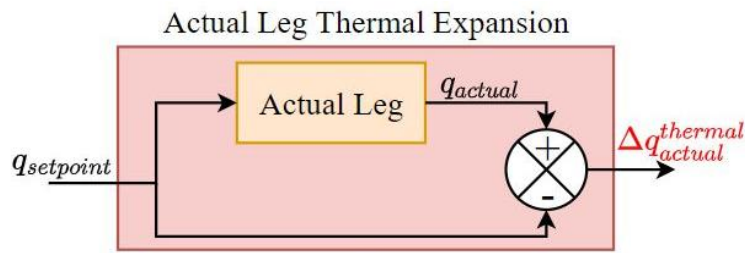

**Figure 2. Actual Leg thermal expansion.** This box compares the requested setpoint to the actual value reached. The "Actual Leg" box represents a position-controlled hexapod leg, wherein a given input setpoint is received and subsequently achieved as position $q_{actual}$.

## 3. Testbed study and design

A dedicated testbed (Fig. 3) was developed to simultaneously measure the temperature profile and thermal expansion of a hexapod leg under heating and cooling cycles, while replicating operational conditions. The leg was mounted vertically to reproduce realistic heat transfer, with joints **(G/M)** preserved to capture inter-articular expansion, but off-axis rotations blocked by flexural elements **(D)**, allowing free axial expansion. To minimize thermal disturbance, these elements were placed away from high-thermal-gradient regions. The extremities were connected to platform pieces **(B)** of the same material as the full hexapod to ensure representativeness. Inter-articular dilatation was monitored during vertical-axis motions using three interferometer heads **(C)** spaced 120° around the top spherical joint, targeting retroreflectors **(F)** near the bottom universal joint. The average of the three beam length variations defined the dilatation. This configuration satisfied the revisited Abbé principle, as defined by Bryan [29], ensuring high accuracy despite small platform rotations. This principle has already been applied in [30] in a similar framework. To reduce refractive air index fluctuations, interferometer beams were enclosed in steel tubes **(E)**, with a weather station used for index correction. Surface temperatures were recorded via thermocouples affixed with aluminum and Kapton tape for measurement reliability. Ambient conditions were stabilized within ±2 K, with variations limited to <1 K/h.

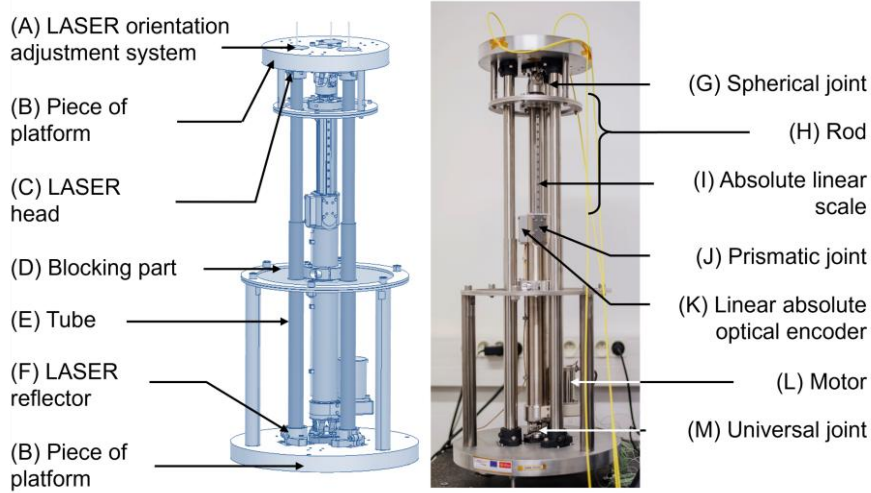

**Figure 3.** Presentation of the complete testbed. The complete CAD of the testbed is shown on the left, while the actual testbed is on the right. The testbed elements are listed.

### 4. A theoretical formula for thermal expansion of the leg

The thermal expansion of an isotropic material is formulated as:
$$\Delta L = \alpha L \Delta T \quad (1)$$
where $\Delta L$ is the length variation and $L$ the initial length. $\alpha$ is the coefficient of linear expansion and $\Delta T$ the temperature change, both assumed homogeneous. For a telescopic leg of the hexapod:
$$L = q_0 + q \quad (2)$$
where $q_0$, $q$ are the contracted inter-articular distance and the relative displacement of the rod. Under homogeneous heating, the modelled leg expansion is given by:
$$\Delta q_{predicted}^{thermal} = A q_0 \Delta T + B q \Delta T \quad (3)$$
where $A q_0 \Delta T$, $B q \Delta T$ denote expansion of the fixed and moving parts, respectively, and $A, B$ are two constants. At this point, the thermal expansion coefficient and the temperature variation are assumed to be homogeneous in each of the two parts of the leg.

In practice, temperature is non-uniform, though preliminary measurements revealed a nearly linear profile along the leg. This suggests that accurate modelling does not require a large number of thermocouples. The number of thermocouples was therefore limited to two per leg. Under this assumption, the expansion model becomes
$$\Delta q_{predicted}^{thermal} = A_i q_0 \Delta T_i + B_i \Delta T_i q + A_j q_0 \Delta T_j + B_j \Delta T_j q \quad (4)$$
where $\Delta T_i$ and $\Delta T_j$ are relative temperature variations at two measurement points, and $A_i, A_j, B_i, B_j$ are constants.

### 5. Estimating the equation parameters

#### *5.1. Measurements*

The constants in (4) were identified experimentally using the testbed, along with the optimal thermocouple locations. A brute-force strategy was adopted: seventeen thermocouples were distributed across the actuator (**Fig. 1**), and all possible pairs were assessed under representative operating scenarios. A simplified case using a single thermocouple was also considered.

User scenarios reflected typical hexapod operations, alternating between "movement" (heating) phases, where the leg followed commanded displacements, and "rest" (cooling) phases, where it maintained position (**Fig. 4**).

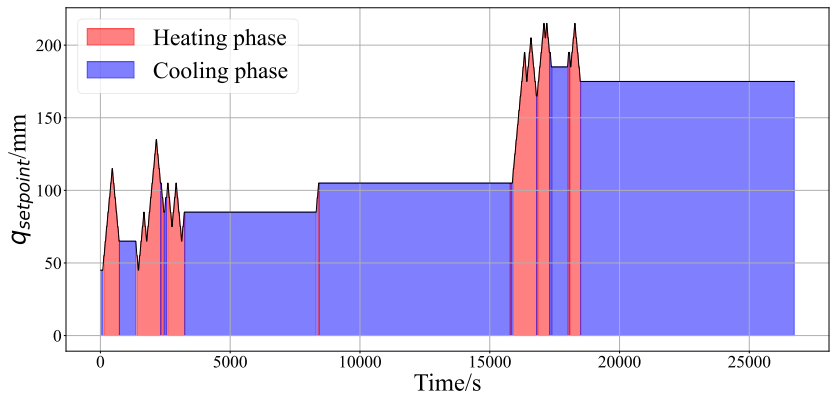

**Figure 4.** Example of a user scenario, the evolution of the setpoint is represented over time.

Interferometric measurements were unreliable during motion and for a few seconds thereafter due to structural vibrations. To address this, short stabilization periods were introduced after each movement. Furthermore, displacements were executed in 10 mm increments to avoid long intervals without valid data. Measurements from all sensors were continuously recorded during both phases. Assuming no thermal expansion at time t=0s, both relative temperatures and interferometric measurements were initialized at this time.

*5.2. Choice of regression method*

An algorithm was applied to the experimental measurements to identify the constants of equation (4) via linear regression, performed separately for each possible thermocouple pairs $i,j$. Here, $\Delta q_{predicted}^{thermal}$ is the response variable, while $q_0 \Delta T_i$, $q_0 \Delta T_j$, $q \Delta T_i$, $q \Delta T_j$ are the explanatory variables (notations summarized in **Fig. 5**).

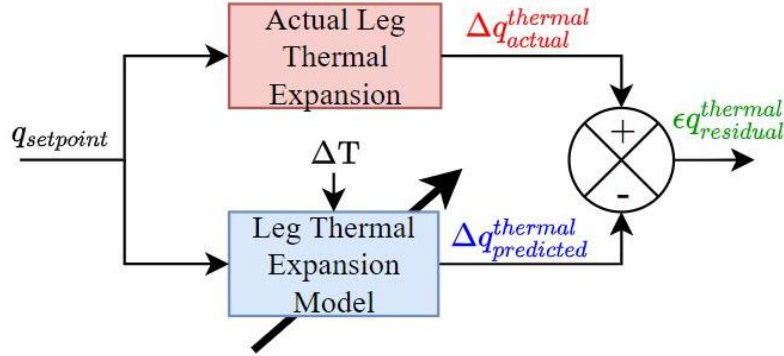

**Figure 5.** Diagram presenting the principle of the thermal expansion prediction model with $\Delta T = [\Delta T_i; \Delta T_j]$. For each couple $\Delta T_i$, $\Delta T_j$, the "Leg Thermal Expansion Model" box is adjusted to minimize $\varepsilon q_{residual}^{thermal}$ according to the chosen regression criterion. The "Actual Leg Thermal Expansion" box presented previously in **Fig. 2** is reused.

The regression cost function was based on the Euclidean norm:

$$L_2 = \sqrt{\sum_{k=1}^{p} \left(\varepsilon q_{k\ residual}^{thermal}\right)^2} \tag{5}$$

with $\varepsilon q_{k\ residual}^{thermal}$ the difference between measured and predicted value of the thermal expansion at time $k$, and $p$ the number of measurements.

*5.3. Classification of thermocouple pairs*

Linear regressions were performed for all thermocouple pairs, as well as for single thermocouples as a subcase. Each regression yielded coefficients for equation (4) that minimized the $L_2$ error between measured and predicted expansions. Thermocouple configurations were then ranked according to their predictive performance. The primary metric was the root mean square error (RMSE), defined as:

$$RMSE = \frac{L_2}{\sqrt{p}} \tag{6}$$

The RMSE, being of the same order as the experimental data, provides a normalized error measure and is widely used for model validation [31]. Outliers, which strongly affect RMSE, were carefully removed following [31]. As a complementary criterion, the infinity norm, $L_\infty$ was introduced.

$$L_\infty = max(|\varepsilon q_{1\ residual}^{thermal}|, \cdots, |\varepsilon q_{p\ residual}^{thermal}|) \tag{7}$$

The results are presented in **Fig. 6**, where each matrix pixel $(i,j)$ corresponds to the performance of the associated couple. The diagonal elements represent single thermocouples, and the matrices are symmetric since equation (4) is invariant under index permutation. The colormap was normalized per matrix, with brightest pixels denoting the best-performing pairs. Several thermocouple pairs could be immediately excluded due to poor performance under either criterion, while the best-performing pairs systematically combined an upper and a lower thermocouple.

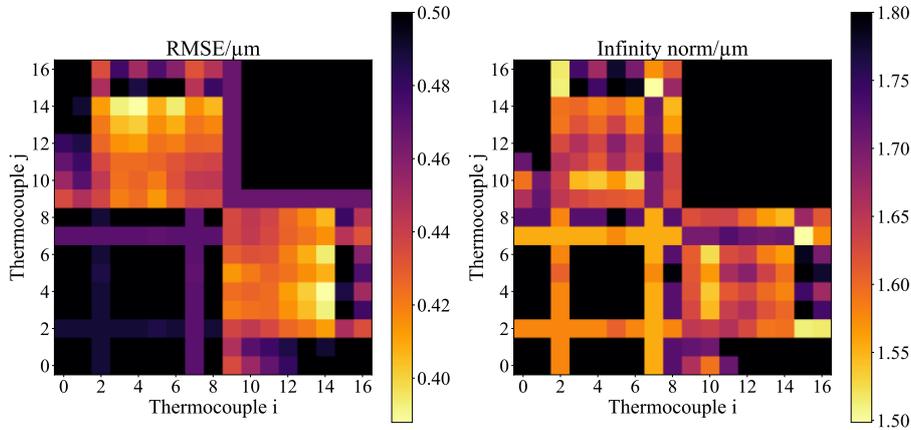

**Figure 6.** Two matrices presenting the values of the two chosen criteria for each pair of thermocouples. The pixel $(i,j)$ of each matrix represents the value of one of the two criteria for the associated couple. The brightest the pixel is, the better the criterion. The single thermocouples are represented on the bottom-left to top-right diagonal, and the matrices are symmetric with respect to that diagonal.

The most efficient configurations are those on the Pareto front. In **Fig. 7** each thermocouple pair is represented according to its coordinates ($RMSE/L_\infty$). Thermocouple pairs (7-15/8-14/6-14/4-14) are the Pareto optima for these two criteria. Thermocouple 7 was the best single thermocouple.

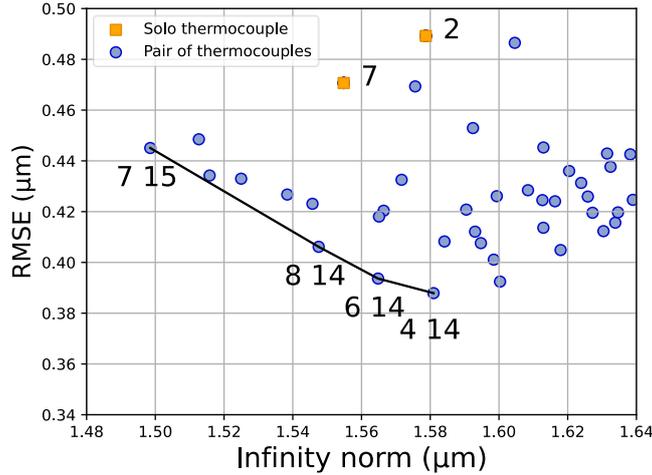

**Figure 7.** Scatter plot, where each thermocouple pair is represented according to its coordinates ($L_\infty$, $RMSE$). Zoom of the zone where the Pareto optima are found, which are linked by a black line. Their names are also given. Single thermocouples are shown in orange.

### 5.4. Verification of the evaluated constants

For cross-validation, the Pareto-optimal pairs and single thermocouple 7 were tested on independent scenarios not used for regression training. The previously identified coefficients were applied to predict thermal expansion, which was then compared against measurements. Figure 8 illustrates five validation tests with thermocouple 7.

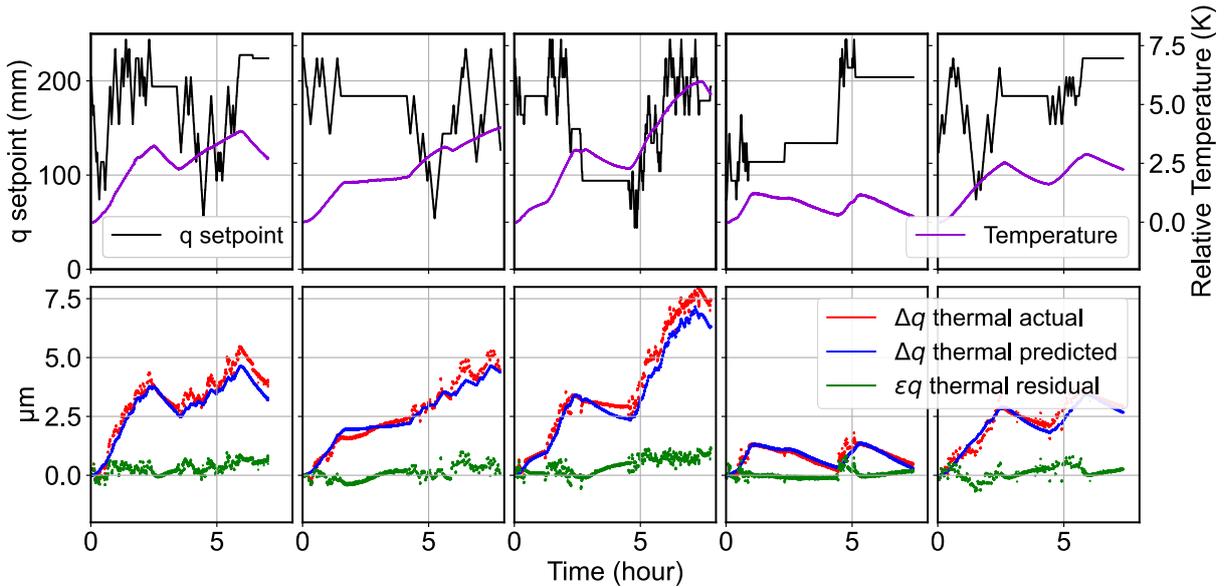

**Figure 8.** Cross-validation results obtained for solo thermocouple 7 for new user scenarios. In the top graph, the black curve depicts leg rod extension, while the violet line indicates the relative temperature of the thermocouple. In the bottom graph, red dots correspond to interferometric measurements of drift, the blue line shows the predicted drift, and the green curve represents the residual deviation between measured and predicted values. All variables were defined in **Fig. 5**.

The maximum thermal drift measured was 7.81 µm, while the maximum residual error between prediction and measurement was limited to 1.28 µm, corresponding to a reduction of over 80%. The mean absolute drift was 2.50 µm, whereas the mean residual error was 0.28 µm, i.e., an 85% reduction. Residual error peaks remained, mostly coinciding with leg displacements, and likely attributable to mechanical defects rather than thermal effects. Tests on the best-performing thermocouple pairs (7–15, 8–14, 6–14, and 4–14) produced results comparable to thermocouple 7 alone, preventing further discrimination among candidates. For practical reasons, only thermocouple 7 was retained. First, its performance was close to the Pareto front: deviations from the best pair were just 0.08 µm ($RMSE$) and 0.06 µm ($L_\infty$), negligible relative to the accuracy target. Second, using a single thermocouple per leg simplifies industrial implementation and reduces integration cost. The fact that one sensor captures the relevant thermal information also confirms that the leg's expansion is well approximated by a low-order thermal gradient.

### 6. Conclusion and perspectives

The final thermal drift model is expressed as:
$$\Delta q_{predicted}^{thermal} = 1.99 \cdot 10^{-3} \, q_0 \Delta T_7 + 1.09 \cdot 10^{-3} q \Delta T_7 \tag{8}$$

with $q$ and $q_0$ being in millimetres, $\Delta q_{predicted}^{thermal}$ in micrometres and $\Delta T_7$ in kelvins. For $q = 0$ (fully retracted leg), the model is similar to the theoretical formula of equation (1). The effective coefficient of thermal expansion of the fixed part is 1.99 µm.m$^{-1}$.K$^{-1}$, consistent with the theoretical coefficient of Invar (1.2–2.0 µm.m$^{-1}$.K$^{-1}$ [32]).

This study developed and validated a regression-based thermal expansion model for a hexapod leg using a dedicated testbed. The approach reduced residual expansion to within 1.5 µm, demonstrating that thermal drift can be effectively compensated by adjusting the commanded position (**Fig. 9**). The model requires only one thermocouple, ensuring minimal hardware complexity and facilitating industrial integration.

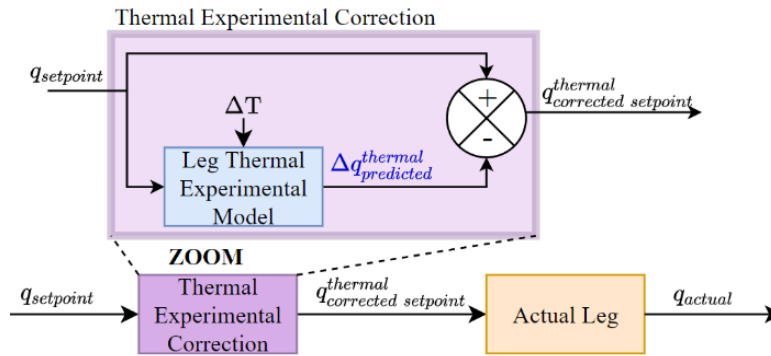

**Figure 9.** Schematic representation of the correction method used. For a requested setpoint, a new thermally corrected setpoint is generated using the leg temperature value.

The methodology can therefore be extended to correct for the thermal expansion of a complete hexapod, either with leg-specific coefficients or, given structural similarity, with a common model. In this way, six-degree-of-freedom compensation could be implemented and validated using a 6D positioning system. Beyond hexapods, the procedure is transferable to other robotic systems: although sensor placement and coefficients will vary, the underlying method remains identical. However, limitations must be acknowledged. The model presumes conduction-dominated heat transfer and has been validated only on metallic materials with conventional thermal properties (e.g., Invar, steel). Applications involving materials of very low conductivity or high expansion may require alternative approaches.

**7. Statements and Declarations**

The authors would like to thank the French Occitanie Region and the European Regional Development Fund (Project FEDER FSE IEJ 2014-2020 LR0019187) for their financial support.